\documentclass[letterpaper]{article}
\usepackage{aaai2027}
\usepackage{microtype}
\usepackage{amsmath,amssymb,amsthm,mathtools}
\usepackage{graphicx,booktabs,array,url}
\usepackage[round,authoryear]{natbib}

\newtheorem{theorem}{Theorem}

\newtheorem{corollary}{Corollary}
\newtheorem{proposition}{Proposition}
\newcommand{\name}{CertBind}
\newcommand{\cmed}{\operatorname{cmed}}
\newcommand{\R}{\mathbb{R}}
\newcommand{\Pp}{\mathbb{P}}
\newcommand{\E}{\mathbb{E}}
\newcommand{\calP}{\mathcal{P}}
\newcommand{\eps}{\varepsilon}
\newcommand{\rank}{\operatorname{rank}}
\newcommand{\spanop}{\operatorname{span}}
\newcommand{\Stab}{\operatorname{Stab}}
\newcommand{\Top}{\operatorname{Top}}
\newcommand{\Proj}{\operatorname{Proj}}

\title{CertBind from Multimodal Connectivity to Certifiable Retrieval Decisions}

\author{
Shuheng Cao$^{1,*}$,
Zhenhao Zhang$^{5,7,*}$,
Ruiqi Chen$^{2,*}$,
Renjie Cao$^{3,\dagger}$,
Weijia Zhang$^{4,\dagger}$,
Siyu Zhang$^{1,\dagger}$,
Jiaxin Liu$^{5,\dagger}$,
Xiangyu Zeng$^{6,\dagger}$,
Haotian Geng$^{7,\dagger}$,
Fan Gu$^{8,\ddagger}$
}

\affiliations{
\begin{tabular}{@{}c@{}}
$^{1}$University of California, San Diego
\quad
$^{2}$University of Michigan, Ann Arbor
\quad
$^{3}$Boston College
\quad
$^{4}$Yale University
\\[0.2em]
$^{5}$ShanghaiTech University
\quad
$^{6}$Nanjing University
\quad
$^{7}$Tsinghua University
\\[0.2em]
$^{8}$Changsha University of Science and Technology
\\[0.45em]
{\small
$^{*}$Co-first authors
\qquad
$^{\dagger}$Equal contribution
\qquad
$^{\ddagger}$Corresponding author
}
\end{tabular}
}

\begin{document}

\maketitle

\begin{abstract}
Lightweight connectors make frozen multimodal encoders composable at the representation level. Deployment exposes a second problem at the level of task decisions. A connected route can expand cross-modal reach while changing an established native retrieval capability. We introduce \name{}, a multiscale theory of certifiable composition for frozen multimodal connector graphs. At the node scale, native anchors establish the exact task identification boundary under the stated chart model. At the edge scale, contract-aware conformal ranks provide graph-wide family-wise error control. At the path scale, an overlap-aware budget and clean calibration yield a finite-sample recovery radius under declared conditions. At the query scale, this radius yields a covered top-$k$ candidate set that becomes a point certificate when its size equals $k$. CertBind therefore retains supported routes as \textsc{Direct}, sends only flagged routes to recovery, returns \textsc{Certified} for decisive recovery, and returns \textsc{Abstain} for unresolved queries. The evaluated C-MCR shared route reduced native CLIP R@1 from $0.524$ to $0.290$. The production fallback recovered $0.963\pm0.002$ of clean retrieval, while the passing branch recorded a no-harm value of $1.000$. CertBind extends multimodal composability from connected representations to certifiable task decisions.
\end{abstract}

\section{Introduction}
Frozen multimodal encoders provide strong but fragmented representations across image, text, audio, and video. Lightweight connector maps make these encoders composable without rebuilding a universal model. Existing systems either synthesize a common space or attach new modalities to a preserved base space \citep{wang2023cmcr,zhang2024exmcr,wang2024freebind,wang2025omnibind}. This progress establishes representation-level composability across frozen spaces and creates multiple routes to downstream tasks.
More broadly, multimodal models are increasingly evaluated by
decision-level capabilities such as explicit affordance reasoning
and action-region grounding \citep{wang2026affordancer1}.

A released connector checkpoint makes the decision-level deployment problem visible. On the same native CLIP image-to-text workload, the evaluated C-MCR shared route reduced R@1 from $0.524$ to $0.290$. A route can expand multimodal reach without becoming the right replacement for an established native route. Connector composition must therefore support expansion and preservation as distinct deployment decisions.

Decision-level certifiability is multiscale because a retrieval output inherits ambiguity from the graph that produced it. At the node scale, graph consistency leaves a residual target gauge relative to a fixed native gallery. At the edge scale, heterogeneous connector contracts require graph-wide calibration. At the path scale, shared failure incidence determines the value of overlapping routes. At the query scale, the observed boundary margin determines whether recovered scores support a point top-$k$ output. Each scale resolves a different ambiguity and supplies the evidence required by the next.

\begin{figure*}[t]
\centering
\includegraphics[width=\textwidth]{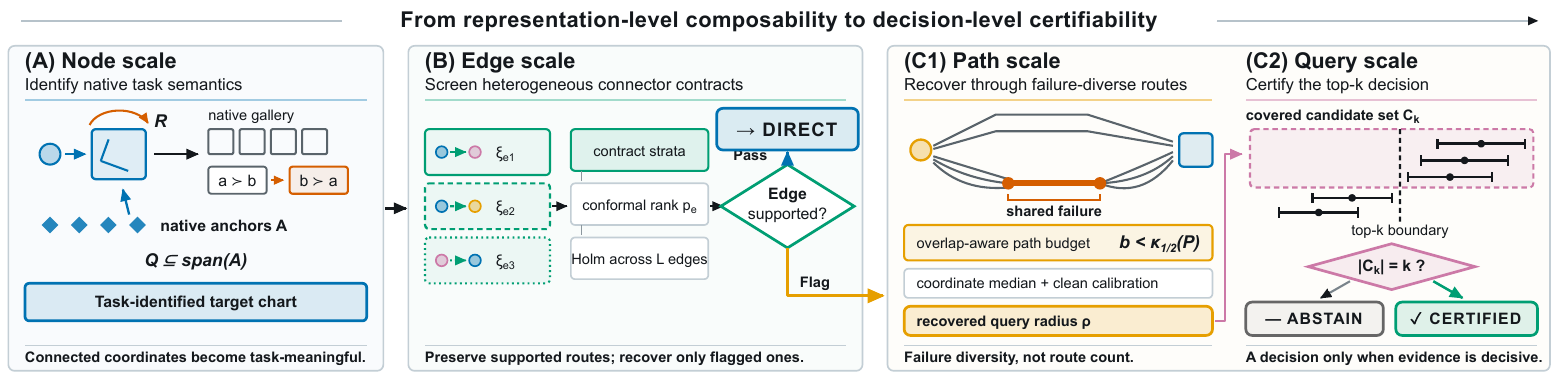}
\caption{CertBind certifies multimodal composition across node, edge, path, and query scales. (A) At the node scale, native anchors fix the target gauge when they span the declared task subspace. (B) At the edge scale, conformal ranks calibrate contract-aware evidence, Holm controls graph-wide decisions, and supported edges retain the registered direct route as \textsc{Direct}. (C) At the path and query scales, flagged edges enter overlap-aware recovery. CertBind returns \textsc{Certified} when the covered top-$k$ candidate set contains $k$ items and \textsc{Abstain} otherwise.}
\label{fig:pipeline}
\end{figure*}

\name{} develops a multiscale theory of certifiable composition from graph structure to retrieval decisions. At the node scale, native anchors identify task-relevant target coordinates. Contract-aware screening determines whether the registered direct edge remains supported. A majority path-transversal budget and clean calibration recover flagged queries through failure-diverse paths. A covered top-$k$ candidate set then certifies decisive outputs and retains set-valued uncertainty otherwise. The resulting preservation-first policy maps the reached certificate level to one deployment action. Supported routes return \textsc{Direct}, flagged routes enter recovery, decisive recovery returns \textsc{Certified}, and unresolved queries return \textsc{Abstain}. Figure~\ref{fig:pipeline} follows this progression from graph structure to the final retrieval decision.

This work makes three contributions to certifiable multimodal composition over frozen connector graphs.
\begin{enumerate}
\item \textbf{Certifiable composition from graph consistency to native task scores.} We separate representation connectivity from native task identification by exposing the residual target gauge. We characterize task-score identification, establish the exact universal anchor rank, and bound registration error under noisy anchors.
\item \textbf{A multiscale certificate chain from edges to queries.} Contract-stratified conformal ranks and Holm control graph-wide clean-edge false flags. A majority path-transversal budget quantifies robustness under shared failures across overlapping routes. Together, these results yield a covered top-$k$ candidate set and an end-to-end error decomposition.
\item \textbf{Preservation-first action and graph design.} A three-state rule maps supported routes, decisive recovery, and unresolved queries to \textsc{Direct}, \textsc{Certified}, and \textsc{Abstain}. The exact path budget is a quota-cover integer program, while edge-disjoint paths yield a polynomial Menger certificate. Released-system cases quantify the operational stakes of preservation, expansion, and recovery.
\end{enumerate}
Full proofs, computational details, and additional deployment records appear in the supplement.

\section{Related Work}
\paragraph{Unified spaces and connector graphs.}
ImageBind, LanguageBind, and UniBind construct broad shared representations \citep{girdhar2023imagebind,zhu2024languagebind,lyu2024unibind}. C-MCR and Ex-MCR connect frozen contrastive spaces, while FreeBind and OmniBind fuse or route among multiple spaces \citep{wang2023cmcr,zhang2024exmcr,wang2024freebind,wang2025omnibind}. Multi-Way Alignment constructs a jointly consistent universe from matched samples \citep{achara2026multiway}. These systems establish representation-level composability and support broad downstream utility. \name{} starts from the resulting frozen graph and studies the task decisions supported after deployment.

\paragraph{Representation identifiability and stitching.}
Linear identifiability, CKA, model stitching, relative representations, and modality-gap analyses study when learned spaces share geometry or communicate \citep{roeder2021identifiability,kornblith2019cka,bansal2021stitching,moschella2023relative,liang2022gap,huh2024platonic}. Together, these analyses establish geometric comparability between learned representation spaces. \name{} specializes the remaining chart ambiguity to native-gallery retrieval, where native anchors determine whether the declared task scores are identified.

\paragraph{Robust recovery and conformal inference.}
Cycle-edge message passing recovers corrupted group-valued relative measurements \citep{lerman2022cemp}. Noisy-correspondence methods repair sample pairs during training \citep{huang2021ncr}. Coordinate medians provide classical clean-majority robustness \citep{huber2009robust}. Conformal ranks provide finite-sample calibration \citep{vovk2005conformal}. Holm's step-down procedure controls family-wise error under arbitrary dependence among valid marginal tests \citep{holm1979}. CertBind carries robust recovery and calibrated uncertainty through shared path failures and observed task margins to a retrieval decision.

Prior work has established multimodal connectivity, representation comparison, robust recovery, and calibrated inference. CertBind develops this decision layer as a multiscale theory linking native task identification, graph-wide screening, overlap-aware recovery, and query-level retrieval certification.

\section{Setting, Threat Model, and Certificate Semantics}
CertBind associates node, edge, path, and query scales with distinct observables, assumptions, and certificates. Routes and calibration roles are fixed before any candidate outcomes are observed.

Let $G=(V,E)$ be a directed connector graph. Node $v$ contains a frozen encoder with space $\mathcal X_v\subseteq\R^{d_v}$. Edge $e=(u,v)$ is a deployed map $T_e:\mathcal X_u\to\mathcal X_v$. The formulation admits nonlinear and noninvertible maps. A declared path family $\calP_{uv}$ provides via outputs $y_p(q)=T_p(q)$ for query $q$. The direct output is $y_0(q)=T_{uv}(q)$.

Before screening, a predeclared anchor operator $\mathcal A_v$ constructs the graph-side matrix $B=\mathcal A_v(A)$ from fixed routes and aggregation rules. The operator and the native anchor matrix $A$ are frozen before candidate outcomes. The identification analysis writes $B=R_*A+E$. Thus $E$ contains the total graph-side anchor discrepancy.

The fitted target registration is also frozen before screening. Henceforth $y_p(q)$ denotes the registered output of route $p$. An edge that passes screening returns its registered direct output without routing-layer modification. A flagged edge is replaced only at the routing layer.

CertBind uses the standard coordinate median and averages the two central order statistics for even sample sizes in each coordinate. The via fallback is
\begin{equation}
\widehat y(q)=\cmed\{y_p(q):p\in\calP_{uv}\}.
\end{equation}
Each edge receives a predeclared audit contract
\begin{equation}
\xi_e=(f_u,f_v,d_u,d_v,\mathsf{map}_e,\mathsf{norm}_e,
\widetilde R_v,\calP_e,D_e,n_e,\nu_e).
\label{eq:contract}
\end{equation}
Its fields specify the encoder families, dimensions, connector class, normalization, frozen registration, route family, score functional, owner count, and owner law. A bounded owner-level score aggregates direct--via disagreement
\begin{equation}
D_e(O_i,\xi_e)\in[0,B],\qquad
A_e=\frac1{n_e}\sum_{i=1}^{n_e}D_e(O_i,\xi_e),
\label{eq:audit}
\end{equation}
where $O_i$ is the statistical owner cluster. Cycle residual and ranking agreement are optional diagnostics.

\paragraph{Four certificate scales.}
A structural observation is invariant along a gauge orbit of graph parameterizations. Node-scale \emph{task identification} asks when every orbit member induces the same native-gallery scores. Edge-scale \emph{screening} controls false connector flags under a calibration law. Path-scale \emph{recovery} combines path incidence and clean-path calibration to bound the recovered target. Query-scale \emph{retrieval certification} guarantees a task decision for a particular recovered query.

\paragraph{Split discipline.}
Anchor owners, screening-control owners, path-calibration owners, and each future query owner occupy separate roles. Conditional on the frozen registered design, screening controls and candidate audit owners are independent of the path-calibration sample and future query owner. Path-calibration owners and the future owner follow the exchangeable deployment law. Strata, routes, and thresholds are fixed independently of candidate outcomes.

Let $Q$ and $G_v$ denote the spans of possible target-space queries and native gallery vectors, and let $W=Q+G_v$. For an unknown target gauge $R$, exact task-score identification requires
\begin{equation}
\Proj_{G_v}(R-I)\Proj_Q=0.
\label{eq:task-specific}
\end{equation}
The universal result below specializes to the common conservative choice $Q=G_v=W$.
All top-$k$ operators use one fixed deterministic rule for score ties.

\begin{table*}[t]
\centering
\scriptsize
\setlength{\tabcolsep}{4.5pt}
\begin{tabular}{p{0.18\textwidth}p{0.22\textwidth}p{0.28\textwidth}p{0.23\textwidth}}
\toprule
Scale & Observable input & Required assumption & Output \\
\midrule
Node identification & native anchor pairs & orthogonal chart model on declared $W$ with anchor rank and conditioning & fixed gauge or impossibility witness \\
Edge screening & owner-level audit scores from clean contract strata & within-stratum exchangeability and optional bad-edge separation & graph-wide false-flag control and miss bound \\
Path recovery & flagged edge and declared path incidence & fewer bad edges than $\kappa_{1/2}$ with calibrated clean-path residuals & radius around the clean target \\
Query certificate & recovered output and native gallery & bounded gallery norms with the calibrated radius event & covered $\mathcal C_k$ and \textsc{Certified} when $|\mathcal C_k|=k$ \\
\bottomrule
\end{tabular}
\caption{Each graph scale resolves an ambiguity left by the preceding scale and produces the evidence required for the next deployment decision.}
\label{tab:layers}
\end{table*}

The threat model separates three adversarial surfaces across the certificate chain. \emph{Gauge ambiguity} changes target coordinates while preserving the graph's structural observations. \emph{Edge corruption} can make every route touching a bad edge arbitrary. \emph{Ordinary approximation error} remains on routes containing no corrupted edge. Under the stated rank conditions, native anchors resolve gauge ambiguity before redundancy is evaluated. Screening power requires failure separation, and graph redundancy operates only after native anchors establish task semantics.

\section{Node-Scale Task Identification from Native Anchors}
At the node scale, certifiable composition requires a target chart that preserves native task scores. A graph may remain internally consistent while the chart rotates relative to a fixed native gallery. Native anchors determine when this residual gauge becomes irrelevant to the declared task.

Although CertBind accepts arbitrary maps, the impossibility result already arises within a restricted submodel. Assume a perfectly consistent target node admits an orthogonal chart and structural observations determine its output only up to $R\in\mathsf O(W)$ on an $r$-dimensional task subspace $W$. Let native anchors be columns of $A=[a_1,\ldots,a_m]$ and define
\begin{equation}
\Stab(A)=\{R\in\mathsf O(W)\mid Ra_i=a_i\ \forall i\}.
\end{equation}

\begin{theorem}[Task identification and anchor complexity]\label{thm:anchor}
Let $S=\spanop(A)$ and $U=S^\perp\cap W$. Under the chart model
\begin{enumerate}
\item native scores for query span $Q$ and gallery span $G_v$ are identifiable if and only if Eq.~\eqref{eq:task-specific} holds for every $R\in\Stab(A)$.
\item $\Stab(A)=I_S\oplus\mathsf O(U)$. Consequently, task scores are identifiable if and only if $Q\subseteq S$ or $G_v\subseteq S$.
\item in the universal case $Q=G_v=W$, identifiability holds if and only if $\rank(A)=r$. When $\rank(A)<r$, a residual reflection preserves every structural observation and anchor but reverses a valid two-item native ranking.
\item suppose $m\ge r$, $\rank(A)=r$, and $\|E\|_{\mathrm{op}}\le\eta_A$. For $\widetilde R=BA^\dagger$
\begin{equation}
\|\widetilde R-R_*\|_{\mathrm{op},W}
\le \frac{\eta_A}{\sigma_r(A)}=: \delta_A.
\label{eq:anchorbound}
\end{equation}
For a unit query observed as $R_*q$, the fixed registration $\widetilde R^\top R_*q$ is within $\delta_A$ of $q$.
\end{enumerate}
\end{theorem}

\paragraph{Proof idea.}
Score invariance is equivalent to
\begin{equation}
\langle(R-I)q,g\rangle=0
\quad\text{for all }q\in Q,\ g\in G_v.
\end{equation}
This is Eq.~\eqref{eq:task-specific}. Every $R\in\Stab(A)$ fixes $S$ pointwise. Orthogonality preserves $U$, and every element of $\mathsf O(U)$ extends to a residual gauge. If either task span lies in $S$, the score change vanishes. Conversely, suppose both spans have nonzero projections onto $U$. Choose $q\in Q$ and $g\in G_v$ with nonzero residual components. An orthogonal action on $U$ can change their inner product. Its identity extension on $S$ changes the task score while fixing every anchor. This proves the finite task-specific criterion. The universal condition follows by taking $Q=G_v=W$.

For noisy anchors, full row rank gives $AA^\dagger=I_W$. Therefore
\begin{equation}
\widetilde R-R_*=(R_*A+E)A^\dagger-R_*=EA^\dagger.
\end{equation}
Submultiplicativity and $\|A^\dagger\|_{\mathrm{op}}=1/\sigma_r(A)$ prove Eq.~\eqref{eq:anchorbound}. Multiplication by the unit vector $R_*q$ gives the registered-query bound. The supplement gives the reflection construction and complete proofs.

\paragraph{Design consequence.}
The theorem gives the exact identification boundary. Task identification depends on anchor coverage of the query or gallery span. Universal identification requires full rank on $W$, while noisy stability depends on anchor conditioning rather than anchor count alone. Under rank deficiency, $\sigma_r(A)=0$ and the universal finite bound is unavailable. At full rank, the stability radius scales as $1/\sigma_r(A)$. A redundant anchor cluster can therefore be less stable than a smaller, well-conditioned spanning set. An offline selector should maximize the smallest singular value on the declared task subspace subject to acquisition cost. If task queries and galleries occupy smaller spans, Eq.~\eqref{eq:task-specific} can certify them without fixing irrelevant target directions.

\begin{proposition}[Optimal bounded-norm anchor geometry]\label{prop:tightframe}
Suppose $m\ge r$ and $\|a_i\|_2\le1$. Then $\sigma_r(A)^2\le m/r$. Equality holds exactly when $AA^\top=(m/r)I_W$ and every anchor has unit norm. For a fixed operator-norm discrepancy budget $\eta_A$, a unit-norm tight frame minimizes the bound in Eq.~\eqref{eq:anchorbound}.
\end{proposition}

\section{Edge-Scale Contract-Aware Graph-Wide Screening}
With node-scale task identification fixed, edge-scale screening asks which registered direct connectors remain supported under their audit contracts. Contract strata define the calibration populations, while graph-wide multiplicity control protects clean registered routes.

Connector edges are rarely exchangeable across all architectures. The audit stratum $s(e)$ is the predeclared equivalence class of the complete contract $\xi_e$. Edges share controls only when every contract field matches. Each control $A_j^{(s)}$ is a complete owner-aggregated score generated under that contract. Write $\mathcal D_s=(A_1^{(s)},\ldots,A_{M_s}^{(s)})$. Define
\begin{equation}
p_s(a)=\frac{1+\sum_{j=1}^{M_s}\mathbf 1\{A_j^{(s)}\ge a\}}{M_s+1},
\qquad p_e=p_s(A_e),
\label{eq:mondrian}
\end{equation}
and, for a frozen batch of $L$ edges,
\begin{equation}
\tau_s=\sup\{a\in[0,B]\mid p_s(a)>\delta_s/L\}.
\label{eq:holm-score-threshold}
\end{equation}
An under-populated stratum returns \textsc{Abstain}.

\begin{theorem}[Stratified conformal screening and graph-wide control]\label{thm:screen}
Suppose a clean candidate score and the clean controls are exchangeable conditional on its predeclared stratum.
\begin{enumerate}
\item $p_e$ is conditionally super-uniform with $\Pp(p_e\le\alpha\mid s(e))\le\alpha$.
\item Holm's procedure at level $\delta_s$ controls the probability of at least one clean-edge false flag by $\delta_s$. Candidate-edge dependence is unrestricted.
\item The smallest attainable $p$-value is $1/(M_s+1)$. Rejection at level $\alpha$ requires $M_s\ge\lceil1/\alpha\rceil-1$.
\item every edge that passes Holm satisfies $p_e>\delta_s/L$ and therefore $A_e\le\tau_s$.
\item conditional on the clean controls, suppose the $n_e$ bad-edge owner scores are independent in $[0,B]$ and satisfy
\begin{equation}
\frac1{n_e}\sum_{i=1}^{n_e}
\E_{\rm bad}[D_e(O_i,\xi_e)\mid\mathcal D_s]
\ge\tau_s+\Delta.
\end{equation}
Then
\begin{equation}
\Pp_{\rm bad}(\mathrm{pass}\mid\mathcal D_s)
\le \exp(-2n_e\Delta^2/B^2).
\label{eq:power}
\end{equation}
\end{enumerate}
\end{theorem}

The first three claims follow from the candidate rank among $M_s+1$ exchangeable scores and the standard Holm step-down argument. For the fourth claim, every unrejected ordered $p$-value exceeds the threshold at the first failed Holm step. That threshold is at least $\delta_s/L$. Hence a pass implies $p_e>\delta_s/L$, which gives $A_e\le\tau_s$ by Eq.~\eqref{eq:holm-score-threshold}.

Conditional on $\mathcal D_s$, define
\[
\mu_{e,s}
:=
\frac{1}{n_e}\sum_{i=1}^{n_e}
\E[D_e(O_i,\xi_e)\mid\mathcal D_s].
\]
Since $\{\mathrm{pass}\}\subseteq\{A_e\le\tau_s\}$, the
mean-separation assumption and Hoeffding's inequality give
\begin{equation}
\begin{aligned}
\Pp_{\rm bad}(\mathrm{pass}\mid\mathcal D_s)
&\le
\Pp_{\rm bad}\bigl(
A_e-\mu_{e,s}\le-\Delta
\mid\mathcal D_s
\bigr)\\
&\le
\exp\!\left(-\frac{2n_e\Delta^2}{B^2}\right).
\end{aligned}
\end{equation}
Thus the screening power statement uses the same pass event as
the deployed Holm procedure.

The screening theorem supplies the edge-scale evidence required by preservation-first routing. Under within-contract exchangeability, it controls graph-wide clean-edge false flags for the frozen batch. Detecting a bad edge additionally requires the stated conditional mean separation. Equation~\eqref{eq:holm-score-threshold} connects that separation directly to the implemented Holm decision. At marginal level $0.05$, rejection requires at least $19$ clean controls. The first Holm threshold for $L$ edges can require substantially more.

Multiple captions, frames, or segments from one item are aggregated before the concentration claim. Holm permits dependence across candidate edges because it needs only marginal super-uniformity.
The family-wise guarantee applies to one frozen deployment batch of $L$ hypotheses. Each later batch requires a new predeclared family with its own error budget.

Screening resolves the edge action by returning supported routes as \textsc{Direct} and sending flagged routes to path-scale recovery under a separate certificate.

\section{Path-Scale Recovery and Query-Scale Retrieval Certification}
At the path scale, a flagged edge activates alternative routes whose value depends on shared failure incidence rather than route count. The certificate measures how many corrupted edges can contaminate a majority of declared paths.

Let $F\subseteq E$ be corrupted edges. A declared path is contaminated when it intersects $F$. For $m=|\calP|$, define
\begin{align}
C_{\calP}(F)&=|\{p\in\calP:p\cap F\ne\varnothing\}|,\\
\kappa_{1/2}(\calP)&=\min_F\{|F|:C_{\calP}(F)\ge\lceil m/2\rceil\}.
\label{eq:kappa}
\end{align}
Overlapping paths earn robustness only when more bad edges are required to contaminate half the route family. For example, ten routes sharing one external edge have $\kappa_{1/2}=1$, whereas five edge-disjoint routes have $\kappa_{1/2}=3$.

Thus $\kappa_{1/2}$ converts shared failure incidence into the strict clean-majority condition required by coordinate-median recovery.

A separate clean calibration split contains native targets $y_i^*$ and the outputs of the same \emph{fixed, anchor-registered} path family used at deployment. Define owner residuals
\begin{equation}
Z_i=\max_{p\in\calP}\|y_{i,p}-y_i^*\|_\infty,
\end{equation}
and let $\widehat\eps_{\delta_p}$ be the split-conformal upper order statistic at rank $\lceil(N+1)(1-\delta_p)\rceil$ (or $+\infty$ when that rank exceeds $N$).

The path budget establishes structural recoverability, while the clean split calibrates a finite-sample radius. The chain then enters the query scale by converting this path evidence into a covered top-$k$ candidate set.

\begin{theorem}[Calibrated sparse-graph prediction set]\label{thm:certificate}
Fix a declared via-edge corruption budget $b<\kappa_{1/2}(\calP)$. Assume future clean-path residuals are exchangeable with the $N$ calibration owners. Also assume the realized corrupted via-edge set satisfies $|F|\le b$. Define
\begin{equation}
\rho:=\sqrt d\,\widehat\eps_{\delta_p}.
\label{eq:rho}
\end{equation}
Under the joint law of the path-calibration sample and one future owner,
\begin{equation}
\Pp(\|\widehat y-y^*\|_2\le\rho)\ge1-\delta_p.
\end{equation}
Let the gallery contain $J$ vectors with norm at most one and fix $1\le k<J$. Write $s_j=\langle\widehat y,g_j\rangle$. Define $\ell_j=s_j-\rho$ and $u_j=s_j+\rho$. Let $\tau_k$ be the $k$th largest lower endpoint and set
\begin{equation}
\mathcal C_k=\{j\mid u_j\ge\tau_k\}.
\label{eq:candidate-set}
\end{equation}
For the clean native set $S_k^*=\Top_k(y^*)$,
\begin{equation}
\Pp(S_k^*\subseteq\mathcal C_k)\ge1-\delta_p.
\label{eq:set-coverage}
\end{equation}
Let $\widehat S_k=\Top_k(\widehat y)$ and let $\widehat\gamma_k$ be its observed boundary gap. Then
\begin{equation}
\lvert\mathcal C_k\rvert=k
\quad\Longleftrightarrow\quad
\widehat\gamma_k>2\rho.
\label{eq:observedmargin}
\end{equation}
Define \textsc{Certified} by this equivalent condition. Then
\begin{equation}
\Pp\bigl(\text{\textsc{Certified}}\cap
\{\widehat S_k\ne S_k^*\}\bigr)
\le\delta_p.
\label{eq:wrong-cert}
\end{equation}
When $\lvert\mathcal C_k\rvert>k$, the method returns $\mathcal C_k$ and abstains from a point top-$k$ output.
\end{theorem}

\paragraph{Proof idea.}
The inequality $|F|<\kappa_{1/2}(\calP)$ leaves a strict majority of $F$-free paths. On the conformal event, every such path lies in the coordinate box centered at $y^*$ with radius $\widehat\eps_{\delta_p}$. Both central order statistics lie in that box in every coordinate. Their coordinate median therefore lies within the radius in Eq.~\eqref{eq:rho}.

For every gallery item, Cauchy--Schwarz gives
\begin{equation}
|\langle\widehat y-y^*,g_j\rangle|\le\rho.
\end{equation}
Hence its clean score lies in $[\ell_j,u_j]$. At least $k$ items have lower endpoint at least $\tau_k$. An item outside $\mathcal C_k$ has upper endpoint below $\tau_k$ and cannot enter $S_k^*$. This proves Eq.~\eqref{eq:set-coverage} on the conformal event.

All intervals have the same radius, so $\tau_k=s_{(k)}-\rho$ and
\begin{equation}
\mathcal C_k=\{j\mid s_j\ge s_{(k)}-2\rho\}.
\end{equation}
It has size $k$ exactly when $s_{(k+1)}<s_{(k)}-2\rho$. This is Eq.~\eqref{eq:observedmargin}. On the coverage event, a certified set contains $S_k^*$ and has the same cardinality. It must equal $S_k^*$. Therefore a certified error can occur only when coverage fails, proving Eq.~\eqref{eq:wrong-cert}. The supplement proves the interval minimality statement.

At the path scale, the theorem maps the budget and calibration to a recovery radius. At the query scale, it maps that radius to a covered top-$k$ candidate set. The output becomes point-valued when the observed margin reduces this set to $k$ items.

The statistical certificate chain leaves two routes to an incorrect non-abstaining output. A separated bad direct edge can pass screening, or a query certificate can fail when its covered radius event fails.

\begin{corollary}[End-to-end error decomposition]\label{cor:end-to-end}
Conditional on the screening controls, consider one bad direct edge that satisfies the separation in Theorem~\ref{thm:screen}. Let the path-calibration sample and future query owner follow the joint law in Theorem~\ref{thm:certificate}, independently of the screening data. Then
\begin{equation}
\Pp(\text{wrong non-abstaining output}\mid\mathcal D_s)
\le e^{-2n_e\Delta^2/B^2}+\delta_p.
\label{eq:end-to-end}
\end{equation}
\end{corollary}

A wrong \textsc{Direct} output requires a bad-edge pass. A wrong \textsc{Certified} output requires failure of the covered radius event. The union bound gives Eq.~\eqref{eq:end-to-end}. Separately, Holm controls the graph-wide probability of changing any clean direct route by $\delta_s$. The corollary closes the statistical chain at the returned query output.

At the path scale, the corruption budget is sharp in the path-output model. An adversary with budget $\kappa_{1/2}$ can choose an edge set that contaminates at least half the paths and can drive the coordinate median unbounded. For cosine retrieval, normalization adds the deterministic factor in the supplement. If $r$ selected paths are edge-disjoint, $\kappa_{1/2}=\lceil r/2\rceil$. Maximizing $r$ gives directed edge connectivity by Menger's theorem \citep{menger1927}. Failure diversity, rather than route count, therefore becomes the next graph-design objective.

\section{Computation, Route Planning, and Preservation-First Decision Rule}
The path budget defines a graph-design objective, while the multiscale chain maps the reached certificate level to one of three actions.

For fixed path-edge incidence $H_{pe}=\mathbf 1\{e\in p\}$ and costs $c_e\ge0$, the weighted exact budget is
\begin{align}
\min_{z,h}\quad&\sum_e c_ez_e\\
\text{s.t.}\quad&h_p\le\sum_eH_{pe}z_e,\quad
\sum_ph_p\ge\lceil m/2\rceil,\\
&z_e,h_p\in\{0,1\}.
\label{eq:ilp}
\end{align}
The weighted explicit-incidence problem is NP-hard by reduction from set cover \citep{karp1972}. Equation~\eqref{eq:ilp} gives an exact offline formulation of this optimization problem. Edge-disjoint routes provide a polynomial-time certificate for deployment planning. A deployment planner should maximize $\kappa_{1/2}$ or its weighted analogue subject to hop, latency, and edge-quality constraints.

\begin{table}[t]
\centering
\scriptsize
\setlength{\tabcolsep}{3.2pt}
\begin{tabular}{p{0.18\columnwidth}p{0.36\columnwidth}p{0.34\columnwidth}}
\toprule
State & Condition & Meaning \\
\midrule
\textsc{Direct} & edge passes Holm-adjusted screen & exact routing-layer preservation after frozen registration \\
\textsc{Certified} & edge flagged, prerequisites hold, and $|\mathcal C_k|=k$ & point top-$k$ output with Eq.~\eqref{eq:wrong-cert} \\
\textsc{Abstain} & $|\mathcal C_k|>k$ or another check is unavailable & covered candidate set $\mathcal C_k$ when available \\
\bottomrule
\end{tabular}
\caption{CertBind maps the certificate level reached by the deployment chain to one of three actions.}
\label{tab:states}
\end{table}

The deployment log records the complete audit contract, anchor rank, $\sigma_r(A)$, conformal $p$-value, Holm threshold, declared budget, path radius, covered top-$k$ candidate set, observed margin, and final state. By construction, the fallback aggregation excludes the registered direct-edge output. Add-only insertion freezes old encoders, maps, registrations, route sets, and thresholds. This rule therefore gives exact zero drift on retained routing layers.

The preservation-first rule operationalizes this chain by retaining supported routes as \textsc{Direct} and sending flagged routes to recovery. Recovery becomes \textsc{Certified} only when the query-scale candidate set contracts to $k$ items. Otherwise, the output is \textsc{Abstain}, with the covered top-$k$ candidate set when available.

\section{Deployment Evidence for Certifiable Composition}
Released checkpoints and recorded protocols expose three operational demands of certifiable composition. Table~\ref{tab:cases} organizes the corresponding evidence as preservation, expansion, and preserve-or-recover behavior.

\begin{table}[t]
\centering
\scriptsize
\setlength{\tabcolsep}{3.0pt}
\begin{tabular}{p{0.18\columnwidth}p{0.35\columnwidth}p{0.35\columnwidth}}
\toprule
Purpose & Recorded comparison & Outcome \\
\midrule
Preservation & C-MCR native route & CLIP I$\to$T R@1 $0.524$ \\
Preservation & C-MCR shared route & CLIP I$\to$T R@1 $0.290$ \\
Expansion & Clotho, C-MCR & T$\to$A R@1 $0.168$ \\
Expansion & Clotho, Ex-MCR & T$\to$A R@1 $0.180$ \\
Expansion & Clotho, towers only & T$\to$A R@1 $0.242$ \\
Expansion & Clotho, graph+Ex & T$\to$A R@1 $0.267$ \\
Routing & Production fallback & clean recovery $0.963\pm0.002$ \\
Routing & Passing branch & measured no-harm $1.000$ \\
\bottomrule
\end{tabular}
\caption{Evidence for preservation, expansion, and preserve-or-recover routing under released checkpoints and recorded protocols.}
\label{tab:cases}
\end{table}

\paragraph{Native-route preservation.}
On the native CLIP image-to-text workload, the evaluated C-MCR shared route changed R@1 from $0.524$ to $0.290$. This comparison identifies native preservation as a separate deployment target.

\paragraph{Cross-modal expansion.}
On the disjoint 245-clip Clotho subset, the original maps were applied without edge refitting. C-MCR and Ex-MCR reached text-to-audio R@1 values of $0.168$ and $0.180$, respectively. Towers alone reached $0.242$, while adding Ex-MCR as optional graph evidence reached $0.267$.

\paragraph{Preserve-or-recover routing.}
Across three seeds, the production via-only fallback recovered $0.963\pm0.002$ of clean retrieval, while the passing branch recorded no-harm $1.000$. A separate five-family blind-median stress suite included the direct output and recovered $0.983$--$0.989$. Together, these cases separate native preservation, cross-modal expansion, and selective recovery.

\section{From Multimodal Connectivity to Certifiable Composition}
Multimodal composability extends beyond representation connectivity to the task scores induced by a composed graph. Graph consistency describes relative structure, while native anchors determine whether that structure preserves fixed native-gallery scores. A connector can therefore remain internally coherent while changing an established retrieval route. This distinction makes task identification the node-scale prerequisite for native preservation.

A related distinction between literal information preservation and
downstream recoverability also appears in lossy text compression,
where strategically deleted content may be reconstructed by an LLM
\citep{zou2026textpreservinglossytextcompression}. CertBind addresses
a different problem: it does not reconstruct omitted content, but
certifies whether connector-induced representation uncertainty
preserves a fixed native retrieval decision.

Calibration is part of composition design because each statistical certificate is indexed by its deployment population. Contract strata determine where edge ranks are valid, while owner units and route families determine path calibration. Broader strata can violate exchangeability, whereas narrower strata reduce attainable $p$-value resolution. Control collection and route design must therefore be planned together.

For median recovery, the relevant unit of graph redundancy is shared failure incidence rather than route count. Paths that share one edge repeat outputs without adding protection against that edge. The majority path-transversal budget converts this failure structure into both a recovery criterion and a route-planning objective.

In CertBind, a recovered representation becomes a certified retrieval output only when its uncertainty clears the query's observed top-$k$ boundary. The calibrated radius induces a covered top-$k$ candidate set whose size depends on the observed margin. Well-separated queries support a point certificate, while boundary-adjacent queries remain set-valued and return \textsc{Abstain}. Certification is therefore a property of the query decision rather than aggregate fallback accuracy.

Task-relevant symmetries provide a route beyond orthogonal charts for certifiable composition. The present identification result studies orthogonal gauges because inner-product retrieval is invariant to a common orthogonal change. Other connector families induce different task-relevant symmetry groups, whose stabilizers determine the corresponding identification condition. Noncompact and nonlinear actions require separate identification and estimation results.

\section{Scope and Ethical Considerations}
The guarantees are indexed by explicit deployment conditions at each certificate scale. The declared anchor condition identifies native task scores under the orthogonal chart model. Within-contract exchangeability establishes clean-edge validity, while independent owner clusters and conditional mean separation give the miss bound. The path theorem permits arbitrary outputs on routes touching corrupted edges when the declared edge budget remains below $\kappa_{1/2}$. Exact budget computation may require offline integer optimization, while edge-disjoint paths provide a polynomial certificate.

The certificate chain also determines the evidence recorded for each deployment decision. Before automatic recovery, an operator declares the task subspace and native gallery. The record identifies native anchors, their conditioning, connector contracts, route incidence, calibration populations, and owner-level path residuals. Any missing required field maps the affected query to \textsc{Abstain}.

CertBind supports auditable deployment decisions and retains human-governed escalation for unresolved queries. Deployment records contain connector and calibration metadata and should receive the same access controls as model artifacts. The reported empirical record covers released checkpoints under the stated protocols.

\section{Conclusion}
Lightweight connector graphs make frozen multimodal systems composable at the representation level. CertBind extends this composability to the decision level through a multiscale certificate chain spanning native task identification, contract-aware graph-wide screening, overlap-aware recovery, and finite-sample top-$k$ certification. An exact node-scale identification boundary, graph-wide edge control, sharp path robustness, and finite-sample query coverage provide the formal basis for this chain. Its preservation-first policy retains supported routes, certifies decisive recovery, and records unresolved queries as \textsc{Abstain}. Connectivity is a property of the graph. Certifiability is a property of the task decision.

\bibliography{references}
\end{document}